%% file: main.tex
\pdfoutput=1

\documentclass{article}
\usepackage{ijcai23}

\usepackage{times}
\usepackage{soul}
\usepackage{url}
\usepackage[hidelinks]{hyperref}
\usepackage[utf8]{inputenc}
\usepackage[small]{caption}
\usepackage{graphicx}
\usepackage{amsmath}
\usepackage{amsthm}
\usepackage{booktabs}
\usepackage{algorithm}
\usepackage{algorithmic}
\usepackage[switch]{lineno}
\usepackage{amssymb} 
\usepackage{mathrsfs}
\usepackage{multirow}
\usepackage{bm}
\usepackage{wrapfig}
\usepackage{subfigure} 

\usepackage[dvipsnames]{xcolor}

\newcommand{\A}{\mathbf{A}}
\newcommand{\I}{\mathbf{I}}

\newtheorem{defn}{Definition}

\newtheorem{theorem}{Theorem}
\newcommand{\sectionname}{Section}
\newcommand{\eqname}{Equation}

\newcommand{\shortname}{LSGNN}
\title{\shortname{}: Towards General Graph Neural Network \\ in Node Classification by Local Similarity}

\author{
Yuhan Chen$^1$\thanks{Equal contribution: Yuhan Chen (draym@qq.com) and Yihong Luo (yluocg@connect.ust.hk).} \ \& \ 
Yihong Luo$^{2,3}$\footnotemark[1] \and
Jing Tang$^{2,3}$\thanks{Correspoding authors: Jing Tang (jingtang@ust.hk) and Chuan Wang (wangchuan@iie.ac.cn).} \and\\
Liang Yang$^4$ \and 
Siya Qiu$^{2,3}$ \and 
Chuan Wang$^5$\footnotemark[2] \and 
Xiaochun Cao$^6$
\affiliations
$^1$School of Computer Science and Engineering, Sun Yat-sen University\\
$^2$The Hong Kong University of Science and Technology (Guangzhou)\\
$^3$The Hong Kong University of Science and Technology\\
$^4$School of Artificial Intelligence, Hebei University of Technology\\
$^5$Institute of Information Engineering, Chinese Academy of Sciences\\
$^6$School of Cyber Science and Technology, Shenzhen Campus of Sun Yat-sen University\\
}
\begin{document}

\maketitle

\input{abstract.tex}
\input{intro.tex}
\input{background.tex}
\input{method.tex}
\input{toy_study.tex}
\input{related_work.tex}
\input{experiment_conclusion}


\section*{ACKNOWLEDGEMENTS}
This work is partially supported 
by the National Key R\&D Program of China under Grant No.\@~2022YFB2703303, 
by the National Natural Science Foundation of China (NSFC) under Grant No.\@~U22B2060, Grant No.\@~61972442, Grant No.\@~62102413, and Grant No.\@~U2001202, 
by Guangzhou Municipal Science and Technology Bureau under Grant No.\@~2023A03J0667, 
by HKUST(GZ) under a Startup Grant, 
by the S\&T Program of Hebei under Grant No.\@~20350802D, 
by the Natural Science Foundation of Hebei Province of China under Grant No.\@~F2020202040, 
and by the Natural Science Foundation of Tianjin of China under Grant No.\@~20JCYBJC00650.

\bibliographystyle{named}
\bibliography{main}

\clearpage
\newpage
\appendix
\input{appendix.tex}


\end{document}

%% file: abstract.tex
\begin{abstract}
Heterophily has been considered as an issue that hurts the performance of Graph Neural Networks (GNNs). To address this issue, some existing work uses a graph-level weighted fusion of the information of multi-hop neighbors to include more nodes with homophily. However, the heterophily might differ among nodes, which requires to consider the local topology. Motivated by it, we propose to use the local similarity (LocalSim) to learn node-level weighted fusion, which can also serve as a plug-and-play module. For better fusion, we propose a novel and efficient Initial Residual Difference Connection (IRDC) to extract more informative multi-hop information. Moreover, we provide theoretical analysis on the effectiveness of LocalSim representing node homophily on synthetic graphs. Extensive evaluations over real benchmark datasets show that our proposed method, namely Local Similarity Graph Neural Network (LSGNN), can offer comparable or superior state-of-the-art performance on both homophilic and heterophilic graphs. Meanwhile, the plug-and-play model can significantly boost the performance of existing GNNs.
Our code is provided at https://github.com/draym28/LSGNN.
\end{abstract}

%% file: intro.tex
\section{Introduction}

Graph Neural Network (GNN) has received significant interest in recent years due to its powerful ability in various real-world applications based on graph-structured data, i.e., node classification~\cite{kipf2016classification}, graph classification~\cite{graphclassification}, and link prediction~\cite{linkprediction}. 
Combining convolutional network and graph signal processing, numerous Graph Convolutional Network (GCN) architectures~\cite{scarselli2008graph,defferrard2016fast,hamilton2017inductive,velivckovic2017attention,kipf2016classification} have been proposed and show superior performance in the above application domains. 
Recent work~\cite{battaglia2018relational} believes that the success of GCN and its variants is built on the homophily assumption: connected nodes tend to have the same class label~\cite{hamilton2020graph}. 
This assumption provides proper inductive bias, raising the general message-passing mechanism: aggregating neighbour information to update ego-node feature\textemdash a special form of low-pass filter~\cite{bo2021beyond}. 

\begin{figure*}[!htbp]
    \centering
    \includegraphics[width=1.0\linewidth]{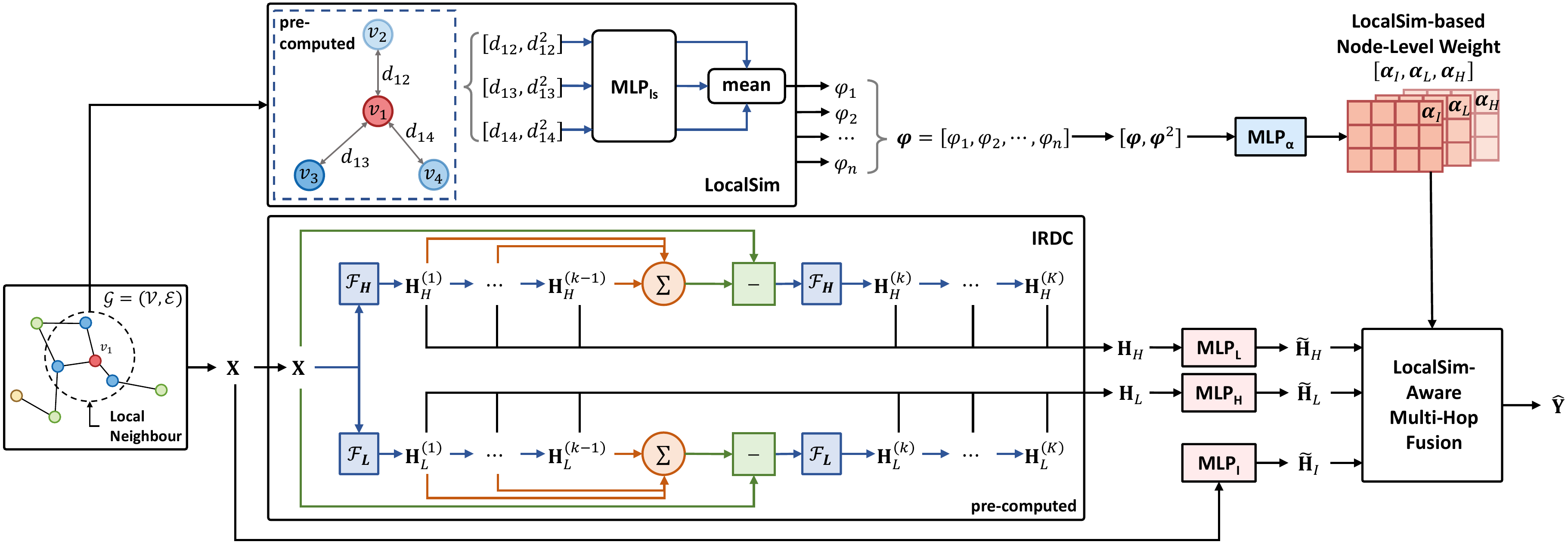}
    \caption{An illustration of \shortname\ framework. `Pre-Computed' means the part can be pre-computed without training.}
    \label{fig:LSGNN_framework}
\end{figure*}

However, the homophily assumption does not always hold~\cite{mcpherson2001birds,jiang2013assortative}. In such cases, empirical evidence shows that GCN may even be worse than simple Multi-Layer Perceptrons (MLPs) that use merely node features as input~\cite{chien2021adaptive} (heterophily issue). A potential explanation is that the low-pass filter is believed to hurt the performance of GCN on heterophilic graph. Recently, the heterophily issue has received the community's interest, and various methods~\cite{pei2020geom,chien2021adaptive,he2021bernnet,li2022finding} have been proposed to address the issue. 
Many existing works~\cite{chien2021adaptive,he2021bernnet,li2022finding} propose to fuse intermediate representations, i.e., outputs of different layers, for heterophilic graphs learning. However, they only consider graph-level heterophily. 
Moreover, intermediate representations extracted by traditional propagation methods~\cite{kipf2016classification} may drop some important information, which limits the effectiveness of information fusion among neighbors. 

We show that in real-world datasets the heterophily among nodes can be significantly different (\sectionname~\ref{motivation}), which suggests that only considering graph-level heterophily is insufficient. 
Although ~\citeauthor{liu2020towards} consider node-level heterophily, they simply map each node's representation to its node-level weight, do not take local topology into consideration, yielding suboptimal results. 
Motivated by the deficiency of existing techniques, we propose to use the local similarity to indicate node homophily to conduct better node-level weighted fusion adaptively. Moreover, we empirically and theoretically show its capabilities on synthetic graphs (\sectionname~\ref{sec: toy study}). 
Furthermore, to obtain more informative intermediate representations for better fusion, we propose a novel Initial Residual Difference Connection (IRDC), which can leverage all the input information. Meanwhile, the IRDC propagation preserves the key property of SGC~\cite{wu2019simplifying} on removing the nonlinear transformations between propagation, which is time- and GPU-consuming in GCNs. This property makes the feature propagation of \shortname\ efficient and deep-learning free. 
Based on the above novel designs, \shortname\ offers high performance and high efficiency in evaluations (\sectionname~\ref{sec: experiment}). 

To summarize, we make the following contributions:
\textbf{1)} We study the local node homophily of real-world datasets and suggest using LocalSim as an indicator of node homophily. Moreover, we empirically and theoretically show the effectiveness of using LocalSim to indicate node homophily on synthetic graphs.
\textbf{2)} We propose \shortname, which contains a LocalSim-aware multi-hop fusion to guide the learning of node-level weight for intermediate representations and an IRDC to extract more informative intermediate representations for better fusion.
\textbf{3)} LocalSim-based node-level weight is a plug-and-play module and can significantly boost the performance of state-of-the-art models, such as H$_2$GCN and GPRGNN.
\textbf{4)} We conduct extensive experiments to demonstrate the superiority of our method, \shortname, which can offer comparable or superior performance against 13 other methods over homophilic and heterophilic graphs.

%% file: background.tex
\section{Preliminaries}
\subsection{Notations}
We denote an undirected graph without self-loops as $\mathcal{G}=(\mathcal{V}, \mathcal{E})$, where $\mathcal{V}=\{v_i\}_{i=1}^n$ is the node set and $\mathcal{E}\subseteq\mathcal{V}\times\mathcal{V}$ is the edge set. Let $\mathcal{N}_i$ denotes the neighbours of node $v_i$. Denote by $\mathbf{A}\in\mathbb{R}^{n\times n}$ the adjacency matrix, and by $\mathbf{D}$ the diagonal matrix standing for the degree matrix such that $\mathbf{D}_{ii}=\sum_{j=1}^n\mathbf{A}_{ij}$. Let $\Tilde{\mathbf{A}}$ and $\Tilde{\mathbf{D}}$ be the corresponding matrices with self-loops, i.e.,~$\Tilde{\mathbf{A}} = \mathbf{A} + \mathbf{I}$ and $\Tilde{\mathbf{D}} = \mathbf{D}+\mathbf{I}$, where $\mathbf{I}$ is the identity matrix. Denote by $\mathbf{X}=\{ \bm{x}_i \}_{i=1}^n\in\mathbb{R}^{n\times d}$ the initial node feature matrix, where $d$ is the number of dimensions, and by $\mathbf{H}^{(k)}=\{\bm{h}_i^{(k)}\}_{i=1}^n$ the representation matrix in the $k$-th layer. We use $\mathbf{Y}=\{ \bm{y}_i \}_{i=1}^n\in\mathbb{R}^{n\times C}$ to denote the ground-truth node label matrix, where $C$ is the number of classes and $\bm{y}_i$ is the one-hot encoding of node $v_i$'s label.

\subsection{Simple Graph Convolution (SGC)}\label{sec:graph_conv}

SGC~\cite{wu2019simplifying} claims that the nonlinearity between propagation in GCN (i.e.,~the ReLU activation function) is not critical, whereas the majority of benefit arises from the propagation itself. Therefore, SGC removes the nonlinear transformation between propagation so that the class prediction $\hat{\mathbf{Y}}$ of a $K$-layer SGC becomes
\begin{equation}
    \hat{\mathbf{Y}} 
    = \operatorname{softmax}\left(
    \mathbf{S}^K \mathbf{X} \mathbf{W}\right), 
    \label{eq:SGC}
\end{equation}
where $\mathbf{S}$ is a GNN filter, e.g.,~$\mathbf{S} = \tilde{\mathbf{D}}^{-\frac{1}{2}}\tilde{\mathbf{A}} \tilde{\mathbf{D}}^{-\frac{1}{2}}$ is used in the vanilla SGC, and $\mathbf{W}$ is the learned weight matrix.  Removing the nonlinearities, such a simplified linear model yields orders of magnitude speedup over GCN while achieving comparable empirical performance. 

\subsection{Heterophily Issue}
Existing GNN models are usually based on the homophily assumption that connected nodes are more likely to belong to the same class. However, many real-world graphs are heterophilic. For example, the matching couples in dating networks are usually heterosexual. Empirical evidence shows that when the homophily assumption is broken, GCN may be even worse than simple Multi-Layer Perceptrons (MLPs) that use merely node features as input~\cite{chien2021adaptive}. Therefore, it is essential to develop a general GNN model for both homophilic and heterophilic graphs.

\citeauthor{pei2020geom}~\shortcite{pei2020geom} introduce a simple index to measure node homophily/heterophily. That is, the homophily $\mathcal{H}(v_i)$ of node $v_i$ is the ratio of the number of $v_i$'s neighbours who have the same labels as $v_i$ to the number of $v_i$'s neighbours, i.e., 
\begin{equation}\label{eq:heterophily}
    \mathcal{H}(v_i) = 
    \frac{| \{ v_j \mid  v_j\in\mathcal{N}_i, \bm{y}_j = \bm{y}_i \} |}{| \mathcal{N}_i |},
\end{equation}
and the homophily in a graph is the average of all nodes, i.e., 
\begin{equation}
    \mathcal{H}(\mathcal{G}) = 
    \frac{1}{n}\sum_{v_i\in\mathcal{V}}\mathcal{H}(v_i).
\end{equation}

Similarly, the heterophily of node $v_i$ is measured by $1-\mathcal{H}(v_i)$, and the heterophily in a graph is $1-\mathcal{H}(\mathcal{G})$.

%% file: method.tex
\section{Methodology}\label{sec: method}
\subsection{Overview}
In this section, we present the \textbf{L}ocal \textbf{S}imilarity \textbf{G}raph \textbf{N}eural \textbf{N}etwork (\shortname), consisting of  a novel propagation method named Initial Residual Difference Connection (IRDC) for better fusion (\textbf{\sectionname~\ref{sec:IRDC-propagation}}), and a LocalSim-aware multi-hop fusion (\textbf{\sectionname~\ref{sec:ls_multichannel_fusion}})  and thus cope with both homophilic and heterophilic graphs. 
We also demonstrate the motivation of using LocalSim as an effective indicator of homophily (\textbf{\sectionname~\ref{motivation}}). \figurename~\ref{fig:LSGNN_framework} shows the framework of \shortname.

\subsection{Initial Residual Difference Connection (IRDC)}\label{sec:IRDC-propagation}
Simply aggregating representation from the previous layer (as applied by most GNNs) might lose important information and consequently incur over-smoothing issues. To obtain more informative intermediate representations, i.e., outputs of different layers, we propose a novel propagation method, namely Initial Residual Difference Connection (IRDC).

\paragraph{General Form.} Inspired by SGC, we remove nonlinearity between layers and formulate IRDC as follows:
\begin{equation}
    \begin{aligned}
        \!\!\!\mathbf{H}^{(1)} &= \operatorname{IRDC}^{(1)}(\mathbf{S},\mathbf{X}) = \mathbf{S} \mathbf{X},\\
        \!\!\!\mathbf{H}^{(k)} &= \operatorname{IRDC}^{(k)}(\mathbf{S},\mathbf{X}) = \mathbf{S} \Big(
        (1-\gamma) \mathbf{X} - \gamma\sum_{\ell=1}^{k-1} \mathbf{H}^{(\ell)}
        \Big),\label{eq:IRDC}
    \end{aligned}
\end{equation}
for $k=2,\dotsc,K$. $\mathbf{S}$ is a GNN filter, and $\gamma\in[0,1]$ is a hyperparameter. The term $\sum_{\ell=1}^{k-1} \mathbf{H}^{(\ell)}$ in \eqname~\eqref{eq:IRDC} can be seen as the total information extracted by the previous $k-1$ layers. In this way, IRDC feeds the information that has never been processed from the original input into the next layer, thus can fully exploit all the information from the initial node features. 
A normalization operation is conducted following IRDC to handle the potential value scale issue. 
See Appendix \ref{app: res_connections_comparison} for comparison between different residual connections. 

\paragraph{Parameterization in Practice.} Many GCNs use adjacency matrix with self-loops as the low-pass GNN filter to extract similar information from neighbours. However, the self-loops added in GCNs may not always be helpful~\cite{zhu2020beyond}. 
Instead, we use enhanced filters~\cite{bo2021beyond} of both low-pass and high-pass ($\mathcal{F}_L+\mathcal{F}_H=\I$) by adding weighted identity matrix $\I$ (i.e., weighted self-loops),
\begin{equation}
    \begin{aligned}
        \mathcal{F}_L&= \beta\mathbf{I} + \mathbf{D}^{-\frac{1}{2}}\mathbf{A}\mathbf{D}^{-\frac{1}{2}}, \\
        \mathcal{F}_H&=\mathbf{I}-\mathcal{F}_L= \left(1-\beta\right)\mathbf{I} - \mathbf{D}^{-\frac{1}{2}}\mathbf{A}\mathbf{D}^{-\frac{1}{2}}, 
    \end{aligned}\label{eq:enhanced_filters}
\end{equation}
where $\beta\in[0,1]$ is a hyperparameter. Intuitively, combining low-pass and high-pass filters together can learn better representations for both homophilic and heterophilic graphs. In a $K$-layer \shortname, the output $\mathbf{H}_L^{(k)}, \mathbf{H}_H^{(k)} \in \mathrm{R}^{n\times d}$ of the $k^{\text{th}}$ low-pass and high-pass graph filter layer are 
\begin{equation}
    \mathbf{H}_{L}^{(k)} = \operatorname{IRDC}^{(k)}(\mathcal{F}_L,\mathbf{X}) 
    \text{ and }
    \mathbf{H}_{H}^{(k)} = \operatorname{IRDC}^{(k)}(\mathcal{F}_H,\mathbf{X}),
    \nonumber
\end{equation}
where $k=1,2,\dotsc,K$. Next, $\mathbf{H}_{L}^{(k)}$ and $\mathbf{H}_{H}^{(k)}$ are transformed into $z$ dimensions by learnable $\mathbf{W}_L^{(k)}, \mathbf{W}_H^{(k)}\in\mathbb{R}^{d\times z}$, i.e.,
\begin{equation}
    \Tilde{\mathbf{H}}_L^{(k)} = \sigma\left( \mathbf{H}_L^{(k)} \mathbf{W}_L^{(k)} \right) \text{ and }
    \Tilde{\mathbf{H}}_H^{(k)} = \sigma\left( \mathbf{H}_H^{(k)} \mathbf{W}_H^{(k)} \right),
    \nonumber
\end{equation}
where $\sigma$ is the ReLU function. To leverage the the initial node features, we also obtain a transformation $\Tilde{\mathbf{H}}_I$ of $\mathbf{X}$ such that
\begin{equation}
    \Tilde{\mathbf{H}}_I = \sigma\left( \mathbf{X} \mathbf{W}_I \right),
    \nonumber
\end{equation}
where $\mathbf{W}_I\in\mathbb{R}^{d\times z}$ is a learned weight matrix. In what follows, $\Tilde{\mathbf{H}}_I, \Tilde{\mathbf{H}}_L^{(k)}, \Tilde{\mathbf{H}}_H^{(k)} \in \mathbb{R}^{n\times z}$ ($k=1,2,\dotsc,K$) are delivered to the LocalSim-aware multi-hop fusion (\sectionname~\ref{sec:ls_multichannel_fusion}) to generate the final node representations.

\subsection{LocalSim: An Indicator of Homophily}\label{motivation}

\begin{figure}[!bpt]
    \centering
    \includegraphics[width=1\linewidth]{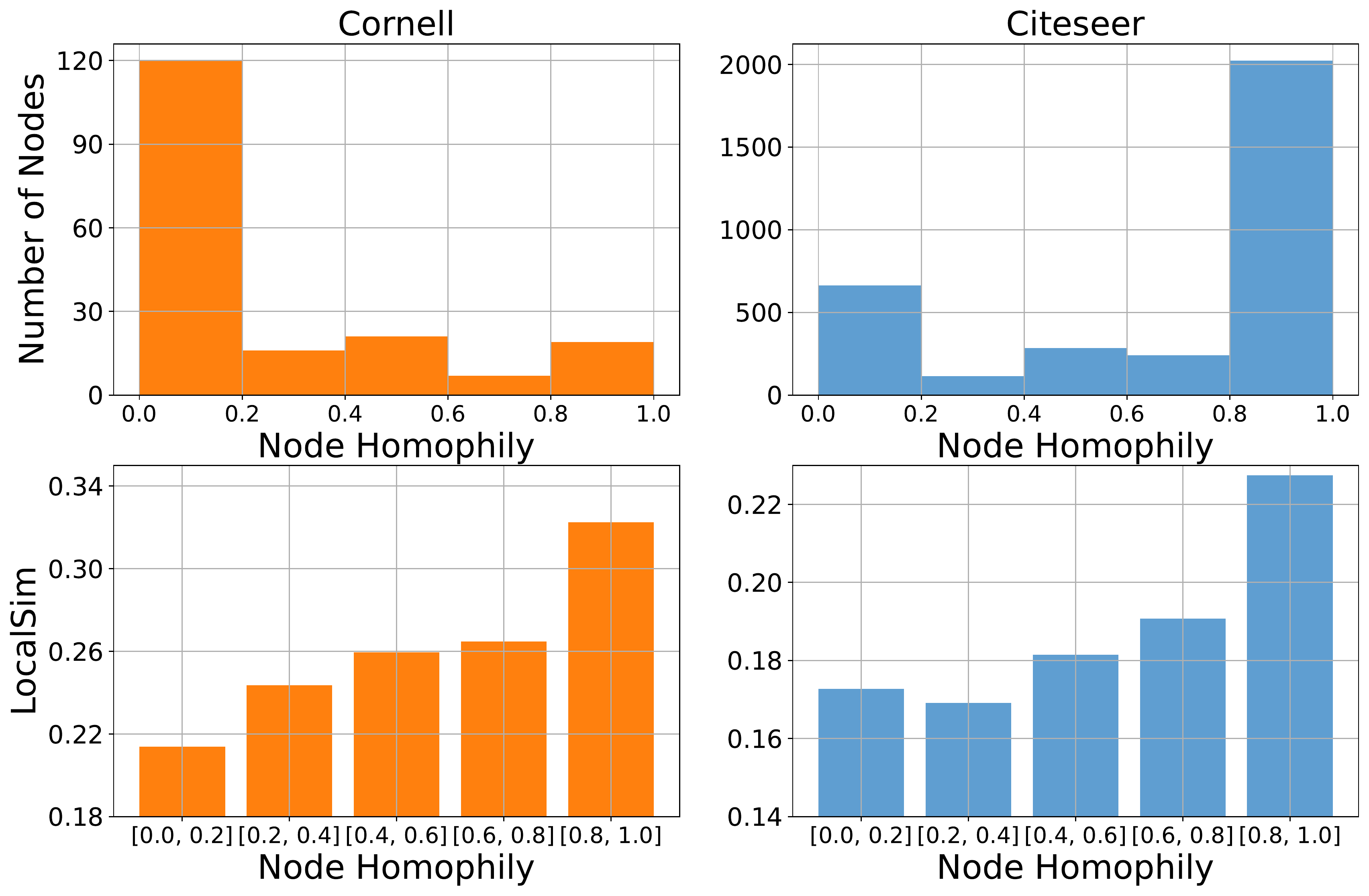}
    \caption{Positive relation between node homophily and LocalSim on Cornell (left column) and Citeseer (right column) datasets. The upper figures show the histogram of node homophily, while the lower figures show the result for node homophily vs.\ LocalSim.}
    \label{fig: homo_ls}
\end{figure}

Based on the hypothesis that nodes with similar features are likely to belong to the same class, we start out by defining a simple version of Local Similarity (LocalSim) $\phi_i$ of node $v_i$ to indicate its homophily, i.e.,
\begin{equation}\label{eq: naive ls}
    \phi_i = \frac{1}{| \mathcal{N}_i |} \sum_{v_j\in\mathcal{N}_i}d_{ij}=\frac{1}{| \mathcal{N}_i |} \sum_{v_j\in\mathcal{N}_i}\operatorname{sim}( \bm{x}_i, \bm{x}_j ),
\end{equation}
where $\operatorname{sim}(\cdot,\cdot):(\mathbb{R}^{d},\mathbb{R}^{d})\mapsto\mathbb{R}$ is a similarity measure, e.g.,
\begin{equation}
    \operatorname{sim}( \bm{x}_i, \bm{x}_j ) = 
    \begin{cases}
        \frac{\bm{x}_i^{\top}\bm{x}_j}{\|\bm{x}_i\|\|\bm{x}_j\|} , &\text{Cosine similarity}, \\
        -\| \bm{x}_i - \bm{x}_j \|_2, &\text{Euclidean similarity}.
    \end{cases}
    \nonumber
\end{equation}
\figurename~\ref{fig: homo_ls} shows the relation between node homophily and LocalSim (using Cosine similarity) on Cornell (left column) and Citeseer (right column) datasets. From the upper figures, we observe that the homophily/heterophily among nodes are different. In particular, Cornell is a heterophilic graph, while Citeseer is a homophilic graph. Moreover, the lower figures show a clear positive correlation between node homophily and LocalSim, which indicates that LocalSim can be used to represent node homophily appropriately.

\subsection{LocalSim-Aware Multi-Hop Fusion}\label{sec:ls_multichannel_fusion}
In \sectionname~\ref{motivation}, we show that LocalSim, leveraging the local topology information, can identify the homophily of nodes. We apply LocalSim to learn the adaptive fusion of intermediate representations for each node. 

The naive LocalSim in \eqname~\eqref{eq: naive ls} simply averages the similarity between the ego-node and its neighbors. In what follows, we refine the definition of LocalSim by incorporating nonlinearity, i.e., $d^2_{ij}$. Specifically, the refined LocalSim $\varphi_i$ of node $i$ is given by
\begin{equation}\label{eq:refined-ls}
    \varphi_i = \frac{1}{| \mathcal{N}_i |} \sum_{v_j\in\mathcal{N}_i} 
    \operatorname{MLP_{ls}}([ d_{ij}, d_{ij}^2 ]),
\end{equation}
where $\operatorname{MLP_{ls}}: \mathbb{R}^2\mapsto\mathbb{R}$ is a 2-layer perceptron. It is trivial to see that incorporating $d^2_{ij}$ via  $\operatorname{MLP_{ls}}$ can encode the variance of series $\{d_{ij}\}_{v_j\in\mathcal{N}_i}$, since $\operatorname{Var}[d_{ij}]=\mathbb{E}[d_{ij}^2] - \mathbb{E}^2[d_{ij}]$ with respect to $v_j\in\mathcal{N}_i$. As a consequence, $\varphi_i$ can better characterize the real ``local similarity'' than $\phi_i$. 
To illustrate, we compare $\varphi_i$ and $\phi_i$ through a simple example. Suppose that node $v_1$ has two neighbours $v_2$ and $v_3$ such that $d_{12}=0$ and $d_{13}=1$, while node $v_4$ has two neighbours $v_5$ and $v_6$ such that $d_{45}=d_{46}=0.5$. Then, $\phi_1=\phi_4=0.5$, which cannot characterize the distinct neighbour distributions. On the other hand, by introducing nonlinearity, we can easily distinguish $\varphi_1$ and $\varphi_4$. Finally, we have the LocalSim vector for all nodes $\bm{\varphi} = [\varphi_1,\varphi_2,\dotsc,\varphi_n] \in \mathbb{R}^n$.

LocalSim can guide the intermediate representations fusion, i.e., using $\bm{\varphi}$ as local topology information when learning weights for fusion. In addition, we also introduce nonlinearity by using $\bm{\varphi}$ and $\bm{\varphi}^2$ together to generate weights, i.e.,
\begin{equation}\label{eq: ls node weight}
    \left[ \bm{\alpha}_I, \bm{\alpha}_L, \bm{\alpha}_H \right] = 
    \operatorname{MLP}_{\alpha}(
    [ \bm{\varphi}, \bm{\varphi}^2 ]
    ), 
\end{equation}
where $\bm{\alpha}_I, \bm{\alpha}_L, \bm{\alpha}_H \in \mathbb{R}^{n\times K}$ are the weights for node representation from each channel and $\operatorname{MLP}_{\alpha}\colon\mathbb{R}^2\mapsto\mathbb{R}^{3K}$ is a 2-layer perceptron. Then, the node representations for the $k$-th layer is computed as follows:
\begin{equation}
    \mathbf{Z}^{(k)} = \bm{\alpha}_I^{(k)}\odot\Tilde{\mathbf{H}}_I
    + \bm{\alpha}_L^{(k)}\odot\Tilde{\mathbf{H}}_L^{(k)}
    + \bm{\alpha}_H^{(k)}\odot\Tilde{\mathbf{H}}_H^{(k)}, 
\end{equation}
where $\mathbf{Z}^{(k)} \in \mathbb{R}^{n\times z}$ and $\bm{\alpha}_I^{(k)}, \bm{\alpha}_L^{(k)}, \bm{\alpha}_H^{(k)} \in \mathbb{R}^{n}$ for $k=1,2,\dotsc, K$, and $\bm{\alpha}_I^{(k)}\odot\Tilde{\mathbf{H}}_I$ is the $i$-th entry of $\bm{\alpha}_I^{(k)}$ times the $i$-th row of $\Tilde{\mathbf{H}}_I$ for $i=1,2,\dotsc,n$ (so do $\bm{\alpha}_L^{(k)}\odot\Tilde{\mathbf{H}}_L^{(k)}$ and $\bm{\alpha}_H^{(k)}\odot\Tilde{\mathbf{H}}_H^{(k)}$). Then, we obtain the final representation as $\Tilde{\mathbf{H}}_I \| \mathbf{Z}^{(1)} \| \mathbf{Z}^{(2)} \| \dotsb \| \mathbf{Z}^{(K)}$, where $\|$ is the concatenation function. Finally, the class probability prediction matrix $\mathbf{\hat{Y}}\in\mathbb{R}^{n\times C}$ is computed by
\begin{equation}
    \mathbf{\hat{Y}} = \left[ \Tilde{\mathbf{H}}_I \| \mathbf{Z}^{(1)} \| \mathbf{Z}^{(2)} \| \dotsb \| \mathbf{Z}^{(K)} \right] \mathbf{W}_{\mathrm{out}}, 
\end{equation}
where $\mathbf{W}_{\text{out}} \in \mathbb{R}^{(K+1)z\times C}$ is a learned matrix.

%% file: toy_study.tex
\section{Case Study of Toy Example}\label{sec: toy study}
In this section, we conduct a case study using a simple synthetic graph to demonstrate the superiority of our LocalSim-based node-level weight.
\subsection{Empirical Investigation} \label{sec: toy empirical investigation}

 Inspired by the Featured Stochastic Block Model (FSBM)~\cite{chanpuriya2022simplified}, in the following, we introduce FSBM with mixture of heterophily that can generate a simple synthetic graph with different node heterophily.

\begin{defn}[FSBM with Mixture of Heterophily]\label{defn:fsbm_ours}
A Stochastic Block Model (SBM) graph $\mathcal{G}$ with mixture of heterophily has $n$ nodes partitioned into $r$ communities $C_1,C_2,\dotsc,C_r$ and consisting of $t$ subgraphs $\mathcal{G}_1,\mathcal{G}_2,\dotsc,\mathcal{G}_t$ where there is no edge between any two subgraphs, with intra-community and inter-community edge probabilities $p_k$ and $q_k$ in subgraph $\mathcal{G}_k$. Let $\bm{c}_1,\bm{c}_2,\dotsc,\bm{c}_r \in \{0,1\}^n$ be indicator vectors for membership in each community, i.e., the $j$-th entry of $\bm{c}_i$ is $1$ if the $j$-th node is in $C_i$ and $0$ otherwise. 
An FSBM is such a graph model $\mathcal{G}$, plus a feature vector $\bm{x} = \bm{f} + \bm{\eta} \in \mathbb{R}^n$, where $\bm{\eta} \sim \mathcal{N}(0, \sigma^2 \mathbf{I})$ is zero-centered, isotropic Gaussian noise and $\bm{f} = \sum_i \mu_i \bm{c}_i$ for some $\mu_1, \mu_2, \dotsc, \mu_r \in \mathbb{R}$, which are the expected feature values of each community.
\end{defn}

According to Definition~\ref{defn:fsbm_ours} and \eqname~\eqref{eq:heterophily}, nodes in the same subgraph have identical homophily in expectation, while nodes from different subgraphs have distinct homophily. We consider FSBMs with $n=1000$, 2 equally-sized communities $C_1$ and $C_2$, 2 equally-sized subgraphs $\mathcal{G}_1$ and $\mathcal{G}_2$, feature means $\mu_1 = 1$, $\mu_2 = -1$, and noise variance $\sigma = 1$. We generate different graphs  by varying $\lambda_1=\frac{p_1}{p_1+q_1}$ and $\lambda_2=\frac{p_2}{p_2+q_2}$, where $\lambda_1,\lambda_2 \in [0,1]$ and high $\lambda_\tau$ means high homophily, with the expected degree of all nodes to 10 (which means $\frac{1}{4}(p_\tau+q_\tau)n=10$ for $\tau=1,2$).

We employ graph-level weight and LocalSim-based node-level weight to classify the nodes. For LocalSim-based node-level weight, we use the proposed method in \sectionname~\ref{sec: method}. For graph-level weight, we replace $\alpha$ in \eqname~\eqref{eq: ls node weight} with learnable graph-level value. For the similarity measure, we use $\operatorname{sim}(x,y) = -(x-y)^2$, where $x$ and $y$ are scalar values. Finally, considering the simplicity of the synthetic networks, we use the simple LocalSim in \eqname~\eqref{eq: naive ls}, and only use adjacency matrix with self-loop $\Tilde{\A} = \A+\I$ and set the layers of IRDC to one.

\begin{figure}[!tbp]
    \centering
    \includegraphics[width=0.8\linewidth]{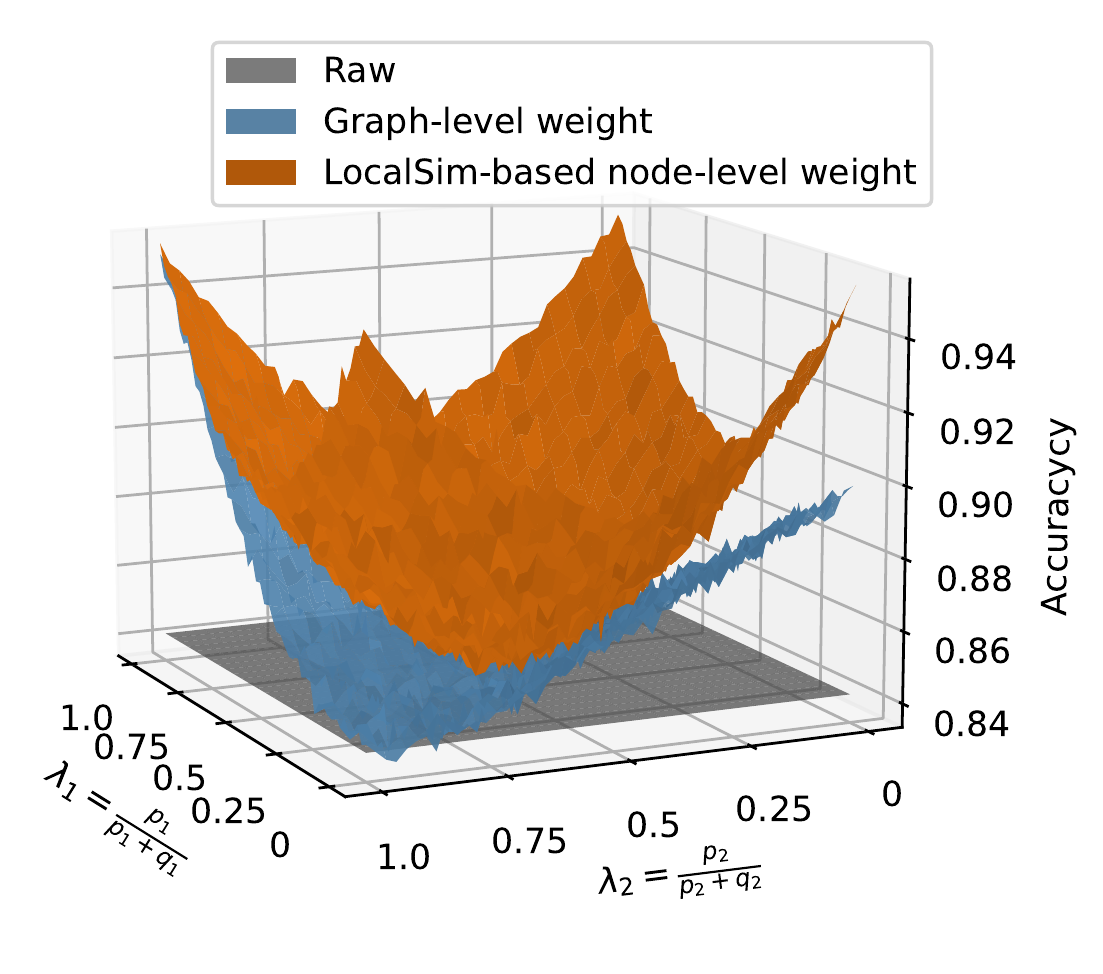}
    \caption{Accuracy (higher is better) on the synthetic graphs using graph-level weight and LocalSim-based node-level weight. Node homophily is measured by $\lambda_i$. `Raw' shows the result directly using raw feature.}
    \label{fig: toy exp}
\end{figure}

As shown in \figurename~\ref{fig: toy exp}, our LocalSim-based node-level weight consistently outperforms graph-level weight. In particular, when node homophily is significantly different among subgraphs, the graph-level weight performs poorly, while our approach still performs very well which can recognize the node homophily well and adaptively provide suitable filters.

\subsection{Theoretical Analysis}
In addition to the empirical investigation, we theoretically confirm the capabilities of LocalSim node-level weight. For simplicity, we assume the number of inter-community and intra-community edges for all the nodes is the same as their expectation and without self-loop. Suppose that the optimal graph-level weight can provide a global filter that achieves the best average accuracy among all subgraphs (not the best in every subgraph), while the optimal node-level weight can provide the optimal filter for each subgraph which can achieve the best accuracy in every subgraph. However, to achieve the optimal node-level weight, we need to estimate the node homophily $\lambda$, and LocalSim is used for the aim. Next, we investigate the effectiveness of using LocalSim to indicate $\lambda$.

\begin{theorem}[Effectiveness of LocalSim Indicating $\lambda$]
\label{homo-ls}
Consider FSBMs with 2 equally-sized communities and 2 equally-sized subgraphs, and assume that the number of inter-community and intra-community edges for all the nodes is the same as their expectation and without self-loop. Consider that LocalSim is defined as $\phi(v_i)=-   \frac{1}{|\mathcal{N}_i |} \sum_{v_j\in\mathcal{N}_i}(x_i-x_j)^2$. For any node $v_i$ that belongs to $\mathcal{G}_\tau$ for $\tau=1,2$, we have $\mathbb{E}[\phi(v_i)] = -2\sigma^2 - (1-\lambda_\tau)(\mu_1-\mu_2)^2$, where $\lambda_\tau = \frac{p_\tau}{p_\tau+q_\tau}$. Moreover, for any two nodes $v_i\in \mathcal{G}_1$ and $v_j\in \mathcal{G}_2$, the expectation of L1-norm distance between $\phi(v_i)$ and $\phi(v_j)$ is no less than to $|\lambda_1-\lambda_2|(\mu_1-\mu_2)^2$.
\end{theorem}

We defer the proof to Appendix \ref{app: proof of homo-ls}. Theorem~\ref{homo-ls} states that $\mathbb{E}[\phi(v_i)]$ linearly correlates to $\lambda_\tau$ when $v_i\in \mathcal{G}_\tau$, which suggests that LocalSim can represent the homophily $\lambda_\tau$ of node $v_i$ unbiasedly. We can also get that the expectation of L1-norm distance positively correlates to $|\lambda_1-\lambda_2|$. This indicates that our estimation is likely to perform better when $|\lambda_1-\lambda_2|$ goes up. When the gap between $\lambda_1$ and $\lambda_2$ shrinks, the two subgraphs become similar to each other, so does the optimal filter for each community. These results confirm the effectiveness of LocalSim indicating node homophily, which yields high performance in our empirical investigation.

%% file: related_work.tex
\section{Additional Related Work}
In this section, we discuss some related work that attempts to address the heterophily issue. Mixhop~\cite{abu2019mixhop} proposes to extract features from multi-hop neighborhoods to get more information. CPGNN~\cite{zhu2020graph} modifies feature propagation based on node classes in GCN to accommodate heterophily. Geom-GCN~\cite{pei2020geom} proposes to use the graph structure defined by geometric relationship to perform aggregation process to address heterophily. FAGCN~\cite{bo2021beyond} proposes to use the attention mechanism to learn the edge-level aggregation weights of low-pass and high-pass filters to adaptively address the heterophily issue, while we use LocalSim to parameterize the node-level weight which achieves less computation complexity and better performance. H$_2$GCN~\cite{zhu2020beyond} and GPRGNN~\cite{chien2021adaptive} propose fusing intermediate representation at graph-level based on their analysis of graphs.  
ACM-GCN~\cite{luan2022revisiting} divides feature propagation into three channels: low-pass, high-pass, and identity, and then adaptively mix the three channels during propagation but does not fuse different intermediate representations.
Compared with H$_2$GCN, GPRGNN, and ACM-GCN, our work only needs propagation once which means the propagation can be pre-computed, while their works are trained via back-propagation through repeated feature propagation which yields huge computation costs. ASGC~\cite{chanpuriya2022simplified} directly uses the least square to filter the features obtained by SGC~\cite{chen2020simple} for classification, while our work involves multiple intermediate information, which is fused by node-level weight. These designs yield better performance based on pre-propagated features. Different from all these methods, our work uses LocalSim to parameterize the node-level weight for better adaptive fusion efficiently, and performs IRDC as an efficient propagation method for better informative representation, thus boosting the performance while keeping high efficiency. Moreover, we show that our designed LocalSim-based node-level weight can be used to improve the performance of H$_2$GCN and GPRGNN, which confirms the superiority of our method.

%% file: experiment_conclusion.tex
\input{main_result.tex}

\section{Experiment} \label{sec: experiment}

\subsection{Datasets and Setups}
We evaluate \shortname\ on nine small real-world benchmark datasets including both homophilic and heterophilic graphs. For homophilic graphs, we adopt three widely used citation networks, Cora, Citeseer, and Pubmed~\cite{PrithvirajSen2008CollectiveCI,LiseGetoor2012QuerydrivenAS}. For heterophilic graphs, Chameleon and Squirrel are page-page Wikipedia networks~\cite{musae}, Actor is a actor network~\cite{pei2020geom}, and Cornell, Texas and Wisconsin are web pages networks~\cite{pei2020geom}. 
Statistics of these datasets can be seen in Appendix \ref{app: dataset stat}.

For all datasets, we use the feature matrix, class labels, and 10 random splits (48\%/32\%/20\%) provided by~\citeauthor{pei2020geom}~\shortcite{pei2020geom}. Meanwhile, we use Optuna to tune the hyperparameter 200 times in experiments. 
For \shortname, we set the number of layer $K=5$. 

There are three types of baseline methods, including non-graph models (MLP), traditional GNNs (GCN, GAT~\cite{velivckovic2017attention}, etc.), and GNNs tolerating heterophily (Geom-GCN, GPRGNN, etc.).

\subsection{Evaluations on Real Benchmark Datasets}
As shown in \tablename~\ref{tab:main_result}, \shortname\ outperforms most of the baseline methods. On homophilic graphs, \shortname\ achieves the SOTA performance on the Cora and Pubmed datasets and is lower than the best result by only 1.28\% on the Citeseer dataset. On heterophilic graphs except Actor, \shortname\ significantly outperforms all the other baseline models by 1.77\%--11.00\%. Specifically, the accuracy of \shortname\ is 7.83\% and 2.97\% higher than GloGNN, the existing SOTA method on heterophilic graphs, on Chameleon and Cornell, is 11.00\% higher than LINKX~\cite{lim2021large} on Squirrel, and is 2.43\% and 1.77\% higher than ACM-GCN on Texas and Wisconsin. 
In particular, \shortname\ achieves the best average accuracy, higher than the second-best by 3.60\%. This demonstrates that \shortname\ can achieve comparable or superior SOTA performance on both homophilic and heterophilic graphs. 

\begin{table}[t]
\centering
\resizebox{\linewidth}{!}
{
\begin{tabular}{c|c|ccc|cc|c}
\toprule
     & $\mathcal{H}(\mathcal{G})$ & GCN & SGC & GCNII & LINKX & GloGNN & \textbf{LSGNN} \\
    \midrule
    arXiv-year & 0.22 & 46.02 & 32.83 & 47.21 & 56.00 & 54.79 & \textbf{56.42} \\
    ogbn-arXiv & 0.66 & 71.74 & 69.39 & \textbf{72.74} & 54.45$^*$ & 49.22$^*$  & 72.64   \\
    Avg & NA & 58.88 & 51.11 & 59.98 & 55.23 & 52.01 & \textbf{64.53}  \\
\bottomrule
\end{tabular}
}
\caption{Average accuracy (\%) over 5 runs on large datasets. The field marked $^*$ denotes our implementation.}
\label{table.results.larger_benchmark}
\end{table}

\subsection{Evaluations on Large Benchmark Datasets}
We also conduct experiments on two large datasets: ogbn-arXiv (homophilic) and arXiv-year (heterophilic), both with 170k nodes and 1M edges but different labels, following settings in \cite{hu2020open} and \cite{lim2021large}, see Appendix \ref{app: dataset stat} for setting details. Table~\ref{table.results.larger_benchmark} shows that our method has competitive performance on both homophilic and heterophilic larger graphs. 

\begin{figure}[t]
    \centering
    \subfigure[Training Cost Time]{
        \includegraphics[width=0.46\linewidth]{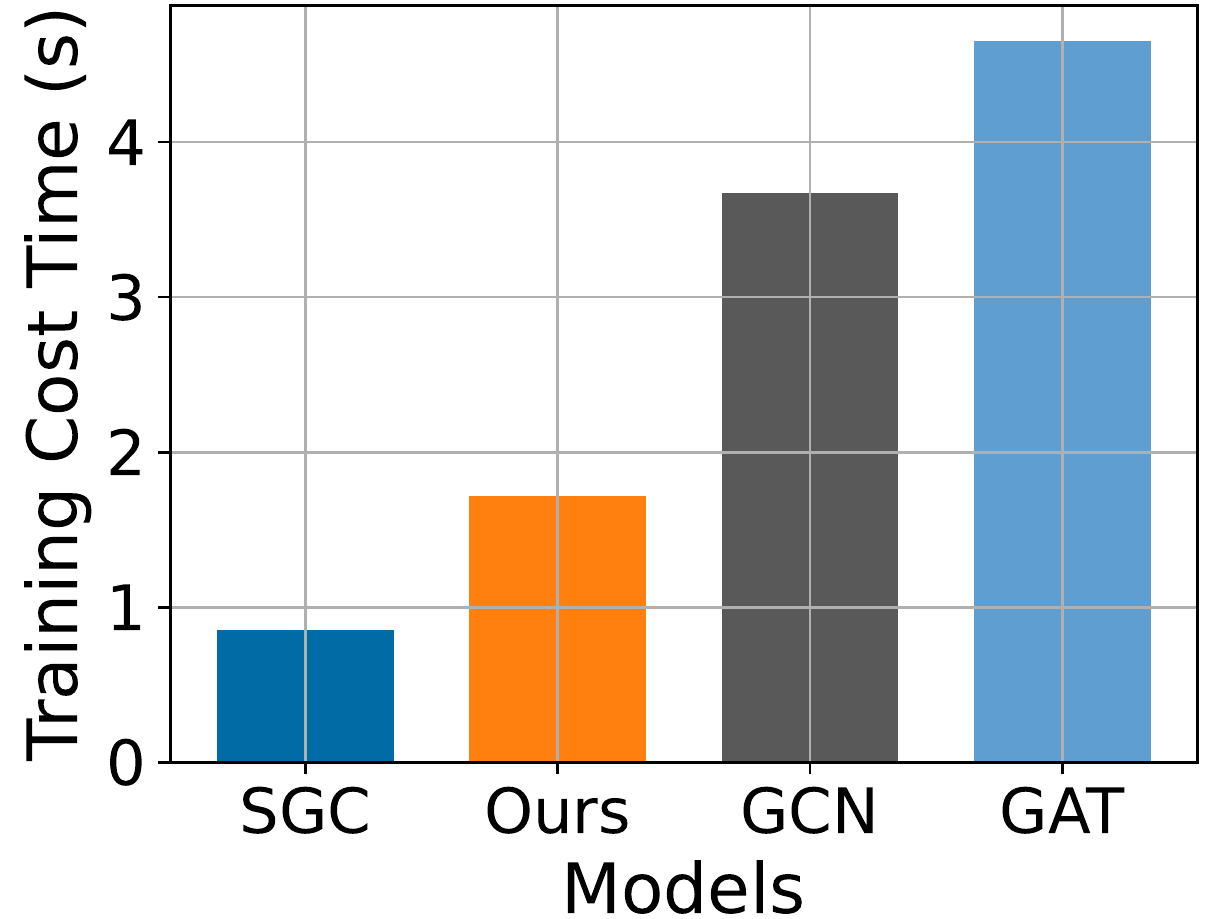}
        \label{fig:comparison on cost time}
    }
	\subfigure[Average Results]{
        \includegraphics[width=0.46\linewidth]{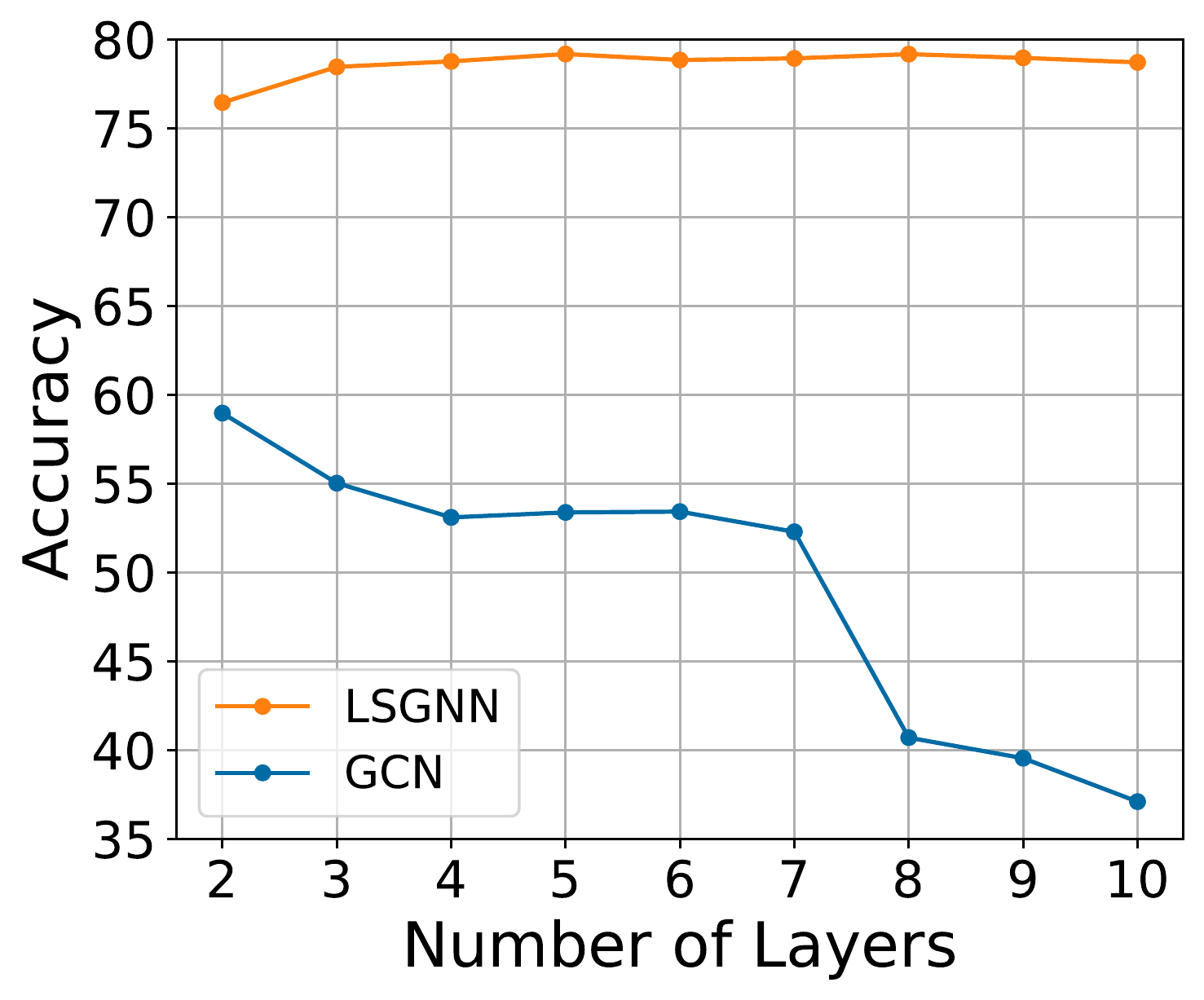}
        \label{fig:oversmooth}
    }
    \caption{Comparison of the average Training Cost Time (s) on nine real-world datasets (Left). Average results with various model depth on nine real-world datasets (Right).}
    \label{over-smoothing}
\end{figure}

\subsection{Training Cost Comparison}

\shortname\ only needs to propagate once during training, and the networks for node-level weight are small. Thus the efficiency of \shortname\ can be close to SGC's. As shown in \figurename~\ref{fig:comparison on cost time}, on the average training time over nine datasets, our proposed method is only slower than SGC, a simple and effective model, and is nearly 2$\times$ faster than GCN and GAT. This confirms the high training efficiency of \shortname. See Appendix \ref{app:comparison on cost time} for more setting details and detailed comparisons of training costs on each dataset.

\subsection{Alleviate Over-Smoothing Issue} \label{sec:over-smooth}
To validate whether \shortname\  can alleviate the over-smoothing issue, we compare the performance between vanilla GCN and \shortname\ under different layers of propagation (\figurename~\ref{fig:oversmooth}), see Appendix \ref{app: detailed_results} for full results. It can be seen that GCN achieves the best performance at 2 layers, and its performance decreases rapidly as layers increase. In contrast, the results of \shortname\ keep stable and high. The reasons are two-fold: (i) our IRDC can extract better informative representations, and (ii) our LocalSim-based fusion can adaptively remain and drop distinguishable and indistinguishable representations, respectively. Through these two designs, \shortname\ can perform well as layers increase, which indicates the capability of \shortname\ to prevent the over-smoothing issue. 

\subsection{LocalSim-based Node-level Weight as a Plug-and-Play Module}
\input{play-and-plug.tex}

\input{ablation_study.tex}

LocalSim-based node-level weight (\sectionname~\ref{sec:ls_multichannel_fusion}) can be regarded as a plug-and-play module, which can be added to existing GNNs that also fuse intermediate representations without involving extra hyperparameters, such as H$_2$GCN, GPRGNN, and DAGNN.


H$_2$GCN and GPRGNN use learned graph-level weight to fuse the intermediate representations, which is only replaced by LocalSim-based node-level weight here. 
Note that we only add LocalSim-based node-level weight into H$_2$GCN and GPRGNN, but not the enhanced filters or the high-pass filters. 
As shown in \tablename~\ref{tab:ls_play-and_plug}, the performance of both H$_2$GCN and GPRGNN increases significantly, indicating that taking each node's local topology into consideration when fusing intermediate representations is more reasonable and effective than merely considering graph-level structure.

DAGNN also fuse intermediate representations by node-level weight, which however is simply mapped from each node's representation and is learned without involving any local topology information, leading to a suboptimal result. We use LocalSim-based node-level weight in DAGNN similarly. The drmatic improving in \tablename~\ref{tab:ls_play-and_plug} suggests that using LocalSim as local topology information can ensure the weight being learned more effectively. Although DAGNN is not designed for heterophilic graphs, it can also perform well on heterophilic graphs after using LocalSim-based node-level weight, and is even superior to H$_2$GCN on Squirrel.

\subsection{Ablation Study}
In what follows, we investigate the effectiveness of LocalSim-based node-level weight, weighted self-loops, and IRDC. 
Specifically, in the baseline model, we use graph-level weight learned without any local topology information to fuse the intermediate representations, add self-loops when conducting graph filtering, and apply the propagation method used in GCNs. Then, the above components are added one by one to the baseline model for performance comparison. As shown in \tablename~\ref{tab:ablation_study}, all the components are effective and can bring great benefits to the model in terms of performance improvement.

Note that learning node-level weight without any local topology information (`Rand') might even bring negative effect on the model (lower than baseline by 0.39\% on average). Once LocalSim is used as local topology information, the model can easily learn an optimal node-level weight (75.16\% on average, higher than the baseline by 1.32\%), which indicates the superiority of LocalSim-Aware Multi-Hop Fusion.

When replacing the fixed self-loops with weighted self-loops, the performance increases significantly (by 2.34\% to 77.50\% on average), especially on heterophilic graphs, such as Chameleon and Squirrel. This is because self-loops might not always be helpful on heterophilic graphs.

IRDC also brings a significant improvement (by an increase of 1.71\% to 79.21\% on average). This suggests that such a propagation method can extract more informative representation for better performance.

\section{Conclusion}
In this paper, \shortname\ is proposed for better performance on both homophilic and heterophilic graphs. 
Many GNNs use graph-level weight to fuse intermediate representations, which does not fully consider the local structure of different nodes. 
Some GNNs use node-level weight learned without involving topology information, yielding suboptimal results. 
Our empirical study and theoretical analysis on synthetic graphs demonstrate the importance of node-level weight considering the local topology information of nodes. The proposed LocalSim-aware multi-hop fusion uses local similarity as guidance to generate a more appropriate node-level weight for intermediate representations fusion, and it can also be used as a plug-and-play module to improve the performance of existing GNNs. 
For better fusion, IRDC is proposed to extract more informative intermediate representations boosting the performance. 
Evaluations over real-world benchmark datasets show the superiority of \shortname\ in handling both homophilic and heterophilic graphs and the effectiveness of all the proposed components.

%% file: main_result.tex
\begin{table*}[t]
    \centering
    \scalebox{0.8}{
    \begin{tabular}{@{}lcccccccccc@{}}
        \toprule
        ~ & \textbf{Cora} & \textbf{Citeseer} & \textbf{Pubmed} & \textbf{Chameleon} & \textbf{Squirrel} & \textbf{Actor} & \textbf{Cornell} & \textbf{Texas} & \textbf{Wisconsin} & \textbf{Avg} \\ 
        \midrule
        $^2$MLP & 74.75 & 72.41 & 86.65 & 46.36 & 29.68 & 35.76 & 81.08 & 81.89 & 85.29 & 65.99  \\ 
        \midrule
        $^2$GCN & 87.28 & 76.68 & 87.38 & 59.82 & 36.89 & 30.26 & 57.03 & 59.46 & 59.80 & 61.62 \\ 
        $^2$GAT & 82.68 & 75.46 & 84.68 & 54.69 & 30.62 & 26.28 & 58.92 & 58.38 & 55.29 & 58.56  \\ 
        $^2$MixHop & 87.61 & 76.26 & 85.31 & 60.50 & 43.80 & 32.22 & 73.51 & 77.84 & 75.88 & 68.10  \\ 
        $^3$GCNII & \underline{88.37} & 77.33 & \underline{90.15} & 63.86 & 38.47 & 37.44 & 77.86 & 77.57 & 80.39 & 70.16  \\ 
        \midrule
        $^1$Geom-GCN & 85.27 & \textbf{77.99} & 90.05 & 60.90 & 38.14 & 31.63 & 60.81 & 67.57 & 64.12 & 64.05  \\ 
        $^1$H$_2$GCN & 87.81 & 77.07 & 89.59 & 59.39 & 37.90 & 35.86 & 82.16 & 84.86 & 86.67 & 71.26  \\ 
        $^1$WRGNN & 88.20 & 76.81 & 88.52 & 65.24 & 48.85 & 36.53 & 81.62 & 83.62 & 86.98 & 72.93  \\ 
        $^3$GPRGNN & 87.95 & 77.13 & 87.54 & 46.58 & 31.61 & 34.63 & 80.27 & 78.38 & 82.94 & 67.45  \\ 
        $^3$LINKX & 84.64 & 73.19 & 87.86 & 68.42 & \underline{61.81} & 36.10 & 77.84 & 74.60 & 75.49 & 71.11  \\ 
        $^3$ACM-GCN & 87.91 & 77.32 & 90.00 & 66.93 & 54.40 & 36.28 & 85.14 & \underline{87.84} & \underline{88.43} & 74.92  \\ 
        $^3$GGCN & 87.95 & 77.14 & 89.15 & 71.14 & 55.17 & \underline{37.54} & 85.68 & 84.86 & 86.86 & 75.05  \\ 
        $^1$GloGNN & 88.33 & \underline{77.41} & 89.62 & \underline{71.21} & 57.88 & \textbf{37.70} & \underline{85.95} & 84.32 & 88.04 & \underline{75.61}  \\ 
        \midrule
        \textbf{LSGNN} & \textbf{88.49} & 76.71 & \textbf{90.23} & \textbf{79.04} & \textbf{72.81} & 36.18 & \textbf{88.92} & \textbf{90.27} & \textbf{90.20} & \textbf{79.21}  \\ 
        \bottomrule
    \end{tabular}}
    \caption{Results on real-world benchmark datasets: Mean accuracy (\%). \textbf{Boldface} and \underline{underline} numbers mark the top-1 and top-2 results, respectively. $^1$ denotes results from their respective papers, $^2$ from the H$_2$GCN paper, and $^3$ from the GloGNN paper.}
    \label{tab:main_result}
\end{table*} 

%% file: play-and-plug.tex
\begin{table*}[!htbp]
    \centering
    \scalebox{0.8}{
    \begin{tabular}{@{}lcccccccccc@{}}
        \toprule
        ~ & \textbf{Cora} & \textbf{Citeseer} & \textbf{Pubmed} & \textbf{Chameleon} & \textbf{Squirrel} & \textbf{Actor} & \textbf{Cornell} & \textbf{Texas} & \textbf{Wisconsin} & \textbf{Avg}  \\ 
        \midrule
        H$_2$GCN & 87.81 & \textbf{77.07} & 89.59 & 59.39 & 37.90 & 35.86 & 82.16 & 84.86 & 86.67 & 71.26  \\ 
        \textbf{w/ LocalSim} & \textbf{88.42} & 76.77 & \textbf{89.91} & \textbf{64.91} & \textbf{41.73} & \textbf{36.29} & \textbf{85.68} & \textbf{85.68} & \textbf{88.82} & \textbf{73.13}  \\ 
        \midrule
        GPRGNN & 87.95 & \textbf{77.13} & 87.54 & 46.58 & 31.61 & 34.63 & 80.27 & 78.38 & 82.94 & 67.45  \\ 
        \textbf{w/ LocalSim} & \textbf{88.83} & 76.60 & \textbf{90.26} & \textbf{65.39} & \textbf{51.96} & \textbf{35.59} & \textbf{85.14} & \textbf{87.84} & \textbf{87.65} & \textbf{74.36}  \\ 
        \midrule
        DAGNN & 82.63 & \textbf{75.47} & 88.50 & 58.03 & 39.59 & 30.20 & 61.89 & 58.92 & 55.10 & 61.15  \\ 
        \textbf{w/ LocalSim} & \textbf{84.12} & 75.29 & \textbf{89.97} & \textbf{61.10} & \textbf{44.91} & \textbf{34.13} & \textbf{79.73} & \textbf{77.03} & \textbf{85.29} & \textbf{70.17}  \\ 
        \bottomrule
    \end{tabular}}
    \caption{LocalSim-based node-level weight as a plug-and-play module added to other models. \textbf{w/ LocalSim} denotes the LocalSim-based node-level weight is used, replacing the graph-level weight used in H$_2$GCN and GPRGNN, and replacing the random node-level weight (learned without any local topology information) in DAGNN. The best results are in \textbf{boldface}.}
    \label{tab:ls_play-and_plug}
\end{table*}

%% file: ablation_study.tex
\begin{table*}[!htbp]
    \centering
    \scalebox{0.75}{
    \begin{tabular}{@{}ccccccccccccccc@{}}
        \toprule
        ~ & \multicolumn{4}{c}{Components} & \multicolumn{9}{c}{Dataset} & ~ \\
        ~ & Rand & LocalSim & w-self-loop & IRDC & \textbf{Cora} & \textbf{Citeseer} & \textbf{Pubmed} & \textbf{Chameleon} & \textbf{Squirrel} & \textbf{Actor} & \textbf{Cornell} & \textbf{Texas} & \textbf{Wisconsin} & \textbf{Avg}  \\ 
        \midrule
        Baseline & ~ & ~ & ~ & ~ & 86.33 & 74.53 & 89.99 & 67.28 & 47.73 & 35.27 & 86.47 & 88.92 & 88.04 & 73.84  \\ 
        ~ & \checkmark & ~ & ~ & ~ & 87.93 & 75.32 & 90.08 & 67.72 & 48.99 & 35.65 & 78.38 & 88.51 & 88.43 & 73.45  \\ 
        ~ & ~ & \checkmark & ~ & ~ & 87.62 & 76.60 & \underline{90.20} & 69.96 & 49.95 & 35.80 & 87.57 & 88.92 & 89.80 & 75.16  \\ 
        ~ & ~ & \checkmark & \checkmark & ~ & 87.65 & \textbf{76.89} & 89.98 & 75.95 & 65.50 & 35.80 & 87.84 & 88.11 & 89.80 & 77.50  \\ 
        \midrule
        \textbf{LSGNN} & ~ & \checkmark & \checkmark & \checkmark & \textbf{88.49} & \underline{76.71} & \textbf{90.23} & \textbf{79.04} & \textbf{72.81} & \textbf{36.18} & \textbf{88.92} & \textbf{90.27} & \textbf{90.20} & \textbf{79.21}  \\ 
        \bottomrule
    \end{tabular}}
    \caption{Ablation study on 9 real-world benchmark datasets. Cells with \checkmark mean that the corresponding components are applied to the baseline model. \textbf{Boldface} and \underline{underlined} mark the top-1 and top-2 results. `Rand' denotes random node-level weight (i.e., learned without any local topology information). `LocalSim' means that the added node-level weight is learned under the guidance of LocalSim. `w-self-loop' denotes weighted self-loops. `IRDC' stands for the propagation via IRDC.}
    \label{tab:ablation_study}
\end{table*}

%% file: appendix.tex
\section{Proof of Theorem 1} \label{app: proof of homo-ls}
For the node $v_i\in \mathcal{G}_\tau$, without loss of generality, suppose that $v_i\in C_1$. (The analysis is analogous when $v_i\in C_2$.) When the number of inter-community and intra-community edges for all the nodes is the same as their expectation and without self-loop, we have
\begin{equation}
    \begin{aligned}
        \mathbb{E}[\phi(v_i)]
        &=-\frac{1}{|\mathcal{N}_i |} \sum_{v_j\in\mathcal{N}_i}\mathbb{E}[(x_i-x_j)^2]\\
        &=-\frac{1}{|\mathcal{N}_i |} \sum_{v_j\in\mathcal{N}_i}\Big(\mathbb{E}[x_i^2]+\mathbb{E}[x_j^2]-2\mathbb{E}[x_ix_j]\Big).
    \end{aligned}
    \nonumber
\end{equation}
By definition, we have
\begin{equation}
    \mathbb{E}[x_i^2]=\operatorname{Var}[x_i]+\mathbb{E}^2[x_i]=\sigma^2+\mu^2_1. 
    \nonumber
\end{equation}
Similarly,
\begin{equation}
    \mathbb{E}[x_j^2]=\operatorname{Var}[x_j]+\mathbb{E}^2[x_j]=
    \begin{cases}
    \sigma^2+\mu^2_1, &\text{if } v_j\in C_1,\\
    \sigma^2+\mu^2_2, &\text{if } v_j\in C_2.
    \end{cases}
    \nonumber
\end{equation}
Moreover, since $v_i$ and $v_j$ are independent random variables, we have
\begin{equation}
    \mathbb{E}[x_ix_j]=\mathbb{E}[x_i]\mathbb{E}[x_j]=
    \begin{cases}
    \mu^2_1, &\text{if } v_j\in C_1,\\
    \mu_1\mu_2, &\text{if } v_j\in C_2.
    \end{cases}
    \nonumber
\end{equation}
Putting it together yields
\begin{equation}
    \begin{aligned}
        \mathbb{E}[\phi(v_i)]
        &=-\frac{1}{|\mathcal{N}_i |} \sum_{v_j\in\mathcal{N}_i\cap C_1}\Big(\sigma^2+\mu^2_1+\sigma^2+\mu^2_1-2\mu^2_1\Big)\\
        &\mathrel{\phantom{=}}-\frac{1}{|\mathcal{N}_i |} \sum_{v_j\in\mathcal{N}_i\cap C_2}\Big(\sigma^2+\mu^2_1+\sigma^2+\mu^2_2-2\mu_1\mu_2\Big)\\
        &=-\frac{p_\tau\cdot 2\sigma^2}{p_\tau+q_\tau}-\frac{q_\tau\cdot (2\sigma^2+(\mu_1-\mu_2)^2)}{p_\tau+q_\tau}\\
        &=-2\sigma^2-(1-\lambda_\tau)(\mu_1-\mu_2)^2.
    \end{aligned}
    \nonumber
\end{equation}
For the second part where $v_i\in \mathcal{G}_1$ and $v_j\in \mathcal{G}_2$, using above result, we have
\begin{equation}
    \begin{aligned}
        \mathbb{E}[\phi(v_i)]&=-2\sigma^2-(1-\lambda_1)(\mu_1-\mu_2)^2,\\
        \mathbb{E}[\phi(v_j)]&=-2\sigma^2-(1-\lambda_2)(\mu_1-\mu_2)^2.
    \end{aligned}
    \nonumber
\end{equation}
Therefore,
\begin{equation}
    \mathbb{E}[|\phi(v_i)-\phi(v_j)|]\geq |\mathbb{E}[\phi(v_i)-\phi(v_j)]|=|\lambda_1-\lambda_2|(\mu_1-\mu_2)^2.
    \nonumber
\end{equation}
This completes the proof.

\section{Comparison Between IRDC and Other Residual Connections}\label{app: res_connections_comparison}
\tablename~\ref{tab:comparison_between_residual_connections} shows the comparison between IRDC and other residuals connections. Note that the coefficient is omitted, $\mathbf{S}$ is a GNN filter, $\mathbf{X}$ is the origin node feature, $\mathbf{H}^{(k)}$ is the intermediate node feature with only propagation, and $\mathbf{Z}^{(k)}$ is the intermediate node feature with both propagation and nonlinear transformation. 

GCN~\cite{kipf2016classification} and SGC~\cite{wu2019simplifying} uses the the output of the previous layer as the input of the next layer, which might result in over-smoothing issue. To overcome this issue, many residual connection methods are proposed. 

GCNII~\cite{chen2020simple} construct a connection to the initial representation, which ensures that the final representation retains at least a fraction of the input
layer even if we stack many layers. 
APPNP~\cite{gasteiger2018predict} apply a similar initial residual connection based on the Personalized PageRank~\cite{page1999pagerank}. 

DRC~\cite{yang2022difference} proposed a novel difference residual connection, enabling each layer to process the information that has not been properly handled in the previous layer. 

Differently, our proposed IRDC aims at enabling each layer to process the total information that has not been handled in all the previous layers, thus can leverage all the input information and extract more informative representations. 

Another key difference between IRDC and other residual connections is that IRDC is pre-computed using only the origin node features, without any trainable parameter or gradient backward propagation, which brings high efficiency. While other residual connections are based on the transformed node representations.

\begin{table}[!ht]
    \centering
    \caption{Comparison between IRDC and other residual connections}
    \scalebox{0.8}{
    \begin{tabular}{@{}ll@{}}
    \toprule
        Residual Connections & Formula \\
    \midrule
        Without Residual Connection (GCN) &  $\mathbf{Z}^{(k)} = \sigma\left(\mathbf{S} \mathbf{Z}^{(k-1)}\mathbf{W}\right)$ \\
        Without Residual Connection (SGC)  &  $\mathbf{Z}^{(k)} = \mathbf{S} \mathbf{Z}^{(k-1)}$ \\
    \midrule
        Initial Residual Connection (GCNII) & $\mathbf{Z}^{(k)} = \sigma\left(\left( \mathbf{Z}^{(0)} + \mathbf{S}\mathbf{Z}^{(k-1)} \right)\mathbf{W}\right)$ \\
        Initial Residual Connection (APPNP) & $\mathbf{Z}^{(k)} = \mathbf{Z}^{(0)} + \mathbf{S}\mathbf{Z}^{(k-1)}$ \\
    \midrule
        Difference Residual Connection (DRC) & $\mathbf{Z}^{(k)} = \mathbf{S} \left( \mathbf{Z}^{(k-2)} - \mathbf{Z}^{(k-1)} \right)$ \\
     \midrule
        \textbf{Initial Residual Difference Connection} & $\mathbf{H}^{(k)} = \mathbf{S} \left( \mathbf{X} - \sum_{\ell=1}^{k-1} \mathbf{H}^{(\ell)} \right)$ \\
    \bottomrule
    \end{tabular}}
    \label{tab:comparison_between_residual_connections}
\end{table}

\section{Comparison on Training Cost Time}\label{app:comparison on cost time}

\paragraph{Experiment setting.}
For GCN, SGC, and GAT, we all use five layers which is the same as our proposed method. And we run the experiments on a computer with an RTX-3080 GPU. We train all models for 200 epochs on all datasets, and we repeat the experiments 10 times and report the average time.

\paragraph{Experiment results.}
The detailed training cost time on each dataset of each model can be found in \tablename~\ref{tab: comparison training cost}. It can be seen that our proposed method is consistently nearly 2$\times$ faster than GCN and GAT, and is close to SGC. 

\section{Additional Experiment Details}\label{app: dataset stat}

\subsection{Statistics Of Benchmark Datasets}
The statistics of nine real-world datasets used in the experiment can be seen in \tablename~\ref{tab:dataset_stat}.

\subsection{Statistics Of Large Benchmark Datasets}
The statistics of two large datasets used in the experiment can be seen in \tablename~\ref{tab:dataset_stat_lsrger_graph}.

\subsection{Experiment Setting Details of Large Benchmark Datasets}

The selected large datasets, ogbn-arXiv and arXiv-year, have same nodes and edges that represent papers and citations. 
In ogbn-arXiv, node labels represent the papers' academic topics, which tend to be a homophily setting. 
While in arXiv-year, the node labels represent the publishing year, thus exhibiting more heterophily label patterns.

For ogbn-arXiv, we use the official split~\cite{hu2020open}, and for arXiv-year, we use the 5 random splits (50\%/25\%/25\%) provided by~\cite{lim2021large}. And note that as observed in SOTA models (e.g., LINKX~\cite{lim2021large} and GloGNN~\cite{li2022finding}) on arxiv-year, incorporating the embedding of the Adjacency matrix is critical for achieving high performance. Therefore, we have also included this embedding in our approach to enhance performance, and details can be seen in our code.

\section{Detailed Results} \label{app:detailed results}

\begin{figure*}
    \centering
    \includegraphics[width=0.8\linewidth]{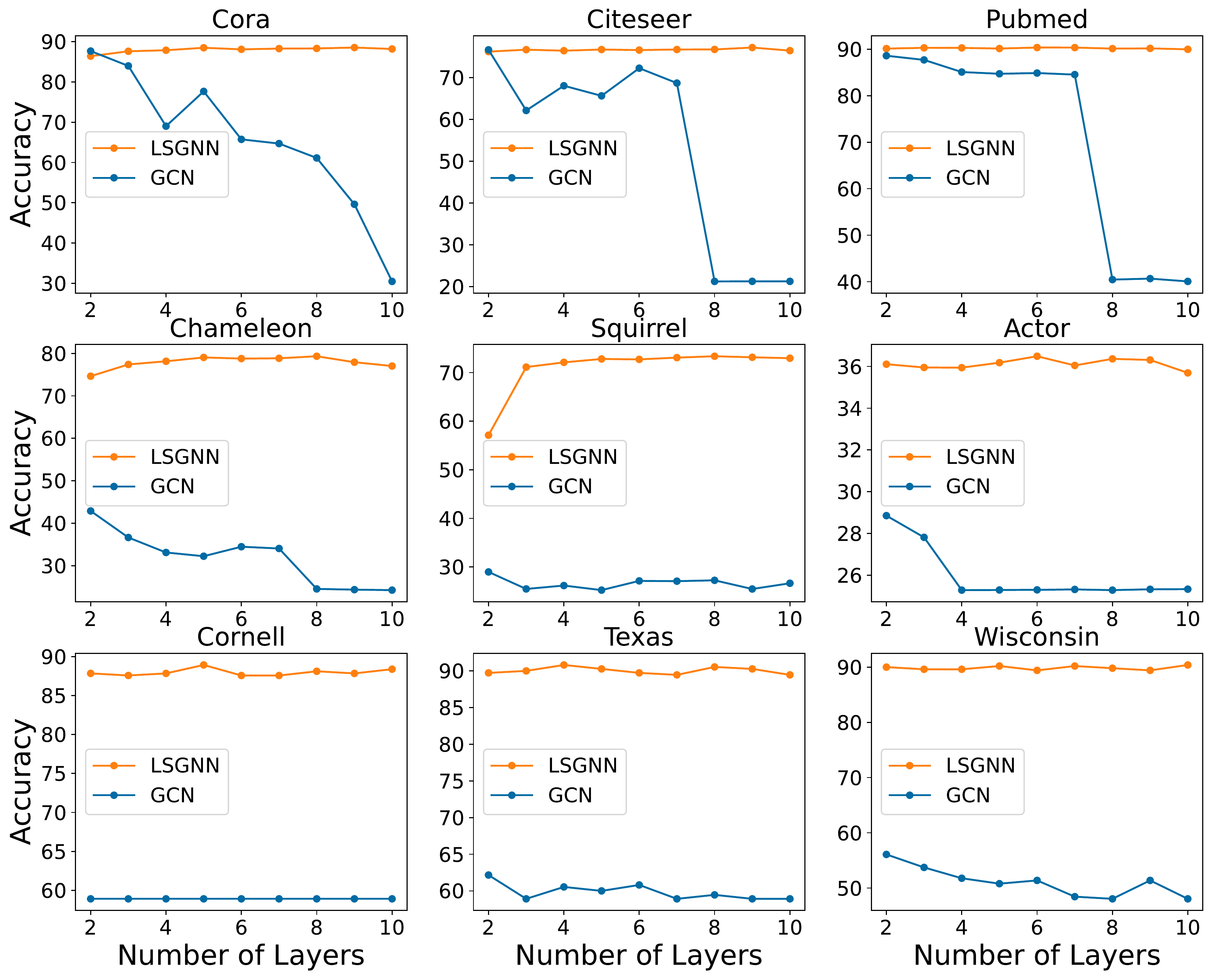}
    \caption{Results with various model depth on nine real-world datasets.}
    \label{fig:oversmooth_full}
\end{figure*}

\paragraph{Full results of Alleviating Over-smoothing Issue in \sectionname~6.4.}
See \figurename~\ref{fig:oversmooth_full} for the full results of comparison on the performance of vanilla GCN and \shortname\  under different layers of propagation. It can be seen that the performance of GCN decreases rapidly as layers increases in most datasets, except Cornell and Squirrel. But we note that Cornell is a simple dataset in which even MLP can achieve $80\%+$ accuracy. And Squirrel only has five classes, while GCN only achieves $25\%\sim30\%$ accuracy which is close to the accuracy of random classification if the class is balanced~($20\%$). In contrast, our method (i.e., LSGNN) achieves high performance on all datasets of all numbers of layers, which indicates the capability of \shortname\ to prevent the over-smoothing issue. Finally, for fairness, we use Optuna to tune the hyper-parameters 200 times for both LSGNN and GCN.

\paragraph{Hyper-parameters Details.}\label{app: detailed_results}
We use Optuna to tune the hyper-parameters 200 times in all experiments. And the number of layers of IRDC is 5 for all experiments, except the over-smoothing experiments. We only set a rough search space of each hyper-parameters for all datasets, as shown in \tablename~\ref{tab:hyper}. For details of the searched hyper-parameters and all other training details, we refer readers to our code which will be released after published.

\input{hyperparam_search_range.tex}
\input{comparison_training_cost.tex}
\input{dataset_stat.tex}
\input{dataset_stat_large_graph.tex}

%% file: hyperparam_search_range.tex
\begin{table}[!htp]
    \centering
    \caption{Hyper-parameters Search Space}
    \label{tab:hyper}
    \scalebox{1}{
    \begin{tabular}{@{}ll@{}}
        \toprule
        \multicolumn{1}{c}{\textbf{Hyper-parameter}} & \multicolumn{1}{c}{\textbf{Search Space}}  \\ 
        \midrule
        learning rate & [1e-3, 1e-1]  \\ 
        weight decay & [1e-6, 1e-1]  \\ 
        dropout & \{0.1, 0.5, 0.6, 0.7, 0.8, 0.9\}  \\ 
        $\beta$ & \{0.1, 0.3, 0.5, 0.7, 0.9, 1.0\}  \\ 
        $\gamma$ & \{0.1, 0.3, 0.5, 0.7, 0.9, 1.0\}  \\ 
        similarity measure & \{cosine, euclidean\}  \\ 
        \bottomrule
    \end{tabular}}
\end{table}

%% file: comparison_training_cost.tex
\begin{table*}[!htbp]
    \centering
    \caption{Time consumption of several GNN models (Seconds/200epoch). These are the time of conducting experiments for 200 epochs on each model, and each experiment is repeated 10 times.} 
    \scalebox{0.9}{
    \begin{tabular}{ccccccccccc}
        \toprule
        ~ & \textbf{Cora} & \textbf{Citeseer} & \textbf{Pubmed} & \textbf{Chameleon} & \textbf{Squirrel} & \textbf{Actor} & \textbf{Cornell} & \textbf{Texas} & \textbf{Wisconsin} & \textbf{Avg} \\ 
        \midrule
        GCN & 3.85 & 3.52 & 3.69 & 3.72 & 3.31 & 3.82 & 3.59 & 3.69 & 3.83 & 3.67  \\ 
        SGC & 0.91 & 0.80 & 1.00 & 0.79 & 0.81 & 0.85 & 0.78 & 0.80 & 0.80 & 0.84  \\ 
        GAT & 4.50 & 4.42 & 5.06 & 4.15 & 6.36 & 4.37 & 4.42 & 4.31 & 4.27 & 4.65  \\ 
        \midrule
        \textbf{LSGNN} & 1.76 & 2.07 & 2.16 & 1.56 & 2.10 & 1.92 & 1.47 & 1.49 & 1.47 & 1.78  \\ 
        \bottomrule
    \end{tabular}}
    \label{tab: comparison training cost}
\end{table*}

%% file: dataset_stat.tex
\begin{table*}[!htbp]
    \centering
    \caption{Benchmark datasets statistics for node classification}
    \label{tab:dataset_stat}
    \scalebox{0.8}{
    \begin{tabular}{@{}cccccccccc@{}}
        \toprule
        ~ & \textbf{Cora} & \textbf{Citeseer} & \textbf{Pubmed} & \textbf{Chameleon} & \textbf{Squirrel} & \textbf{Actor} & \textbf{Cornell} & \textbf{Texas} & \textbf{Wisconsin} \\ 
        \midrule
        \# Nodes & 2708 & 3327 & 19717 & 2277 & 5201 & 7600 & 183 & 183 & 251 \\ 
        \# Edges & 5278 & 4552 & 44324 & 18050 & 108536 & 15009 & 149 & 162 & 257 \\ 
        \# Classes & 7 & 6 & 3 & 5 & 5 & 5 & 5 & 5 & 5  \\ 
        \# Features & 1433 & 3703 & 500 & 2325 & 2089 & 932 & 1703 & 1703 & 1703 \\ 
        $\mathcal{H}(\mathcal{G})$ & 0.81 & 0.74 & 0.80 & 0.24 & 0.22 & 0.22 & 0.31 & 0.11 & 0.20 \\
        \bottomrule
    \end{tabular}}
\end{table*}

%% file: dataset_stat_large_graph.tex
\begin{table*}[!htbp]
    \centering
    \caption{Large Benchmark datasets statistics for node classification}
    \label{tab:dataset_stat_lsrger_graph}
    \scalebox{0.8}{
    \begin{tabular}{@{}cccccc@{}}
        \toprule
        ~ & \#Nodes & \# Edges & \# Classes & \# Features & $\mathcal{H}(\mathcal{G})$  \\ 
        \midrule
        ogbn-arXiv & 169343 & 1166243 & 40 & 128 & 0.66  \\ 
        arXiv-year & 169343 & 1166243 & 5 & 128 & 0.22  \\ 
        \bottomrule
    \end{tabular}}
\end{table*}